\documentclass[journal]{IEEEtran}
\ifCLASSINFOpdf
\else
\fi
\usepackage{cite}
\usepackage{graphicx}
\usepackage{caption}
\usepackage{subcaption}

\usepackage{color}

\begin{document}
\title{Photo Filter Recommendation\\by Category-Aware Aesthetic Learning}

\author{Wei-Tse Sun, Ting-Hsuan Chao, Yin-Hsi Kuo, Winston H. Hsu
\thanks{W.-T. Sun and T.-H. Chao are with the Department of Computer Science and Information Engineering, National Taiwan University, Taipei 10617, Taiwan (e-mail: r03922071@ntu.edu.tw; r02922047@ntu.edu.tw). }
\thanks{Y.-H. Kuo is with the Graduate Institute of Networking and Multimedia, National Taiwan University, Taipei 10617, Taiwan (e-mail: kuonini@cmlab.csie.ntu.edu.tw). }
\thanks{W. H. Hsu is with the Graduate Institute of Networking and Multimedia and the Department of Computer Science and Information Engineering, National Taiwan University, Taipei 10617, Taiwan (e-mail: whsu@ntu.edu.tw). }}


\markboth{IEEE TRANSACTION ON MULTIMEDIA}%
{Photo Filter Recommendation by Category-Aware Aesthetic Learning}



\maketitle

\begin{abstract}
Nowadays, social media has become a popular platform for the public to share photos. To make photos more visually appealing, users usually apply filters on their photos without domain knowledge. However, due to the growing number of filter types, it becomes a major issue for users to choose the best filter type. For this purpose, filter recommendation for photo aesthetics takes an important role in image quality ranking problems. In these years, several works have declared that Convolutional Neural Networks (CNNs) outperform traditional methods in image aesthetic categorization, which classifies images into high or low quality. Most of them do not consider the effect on filtered images; hence, we propose a novel image aesthetic learning for filter recommendation. Instead of binarizing image quality, we adjust the state-of-the-art CNN architectures and design a pairwise loss function to learn the embedded aesthetic responses in hidden layers for filtered images. Based on our pilot study, we observe image categories (e.g., portrait, landscape, food) will affect user preference on filter selection. We further integrate category classification into our proposed aesthetic-oriented models. To the best of our knowledge, there is no public dataset for aesthetic judgment with filtered images. We create a new dataset called Filter Aesthetic Comparison Dataset (FACD). It contains 28,160 filtered images based on the AVA dataset and 42,240 reliable image pairs with aesthetic annotations using Amazon Mechanical Turk. It is the first dataset containing filtered images and user preference labels. We conduct experiments on the collected FACD for filter recommendation, and the results show that our proposed category-aware aesthetic learning outperforms aesthetic classification methods (e.g., 12\% relative improvement). 
\end{abstract}

% most social media provide filters by which users can change the appearance of their photos without domain knowledge
% pairwise ranking loss 
% traditional aesthetic classification
% a novel method for image aesthetic learning
%  By utilizing pairwise image comparison, the models embed aesthetic responses in the hidden layers.
% To improve the filter aesthetic ranking,
% users usually prefer different filters for various image categories (e.g., portrait, landscape, food)
% aesthetic comparison annotations 
% Based on the collected FACD, we conduct experiments on our proposed methods for photo filter recommendation. Experimental results show that

\begin{IEEEkeywords}
Convolutional Neural Network, Filter Recommendation, Image Quality, Aesthetic, Pairwise Comparison. 
\end{IEEEkeywords}

% \fi
\IEEEpeerreviewmaketitle

\section{Introduction}

\begin{figure}[t]
\centering
\includegraphics[width=3.5in, keepaspectratio=true]{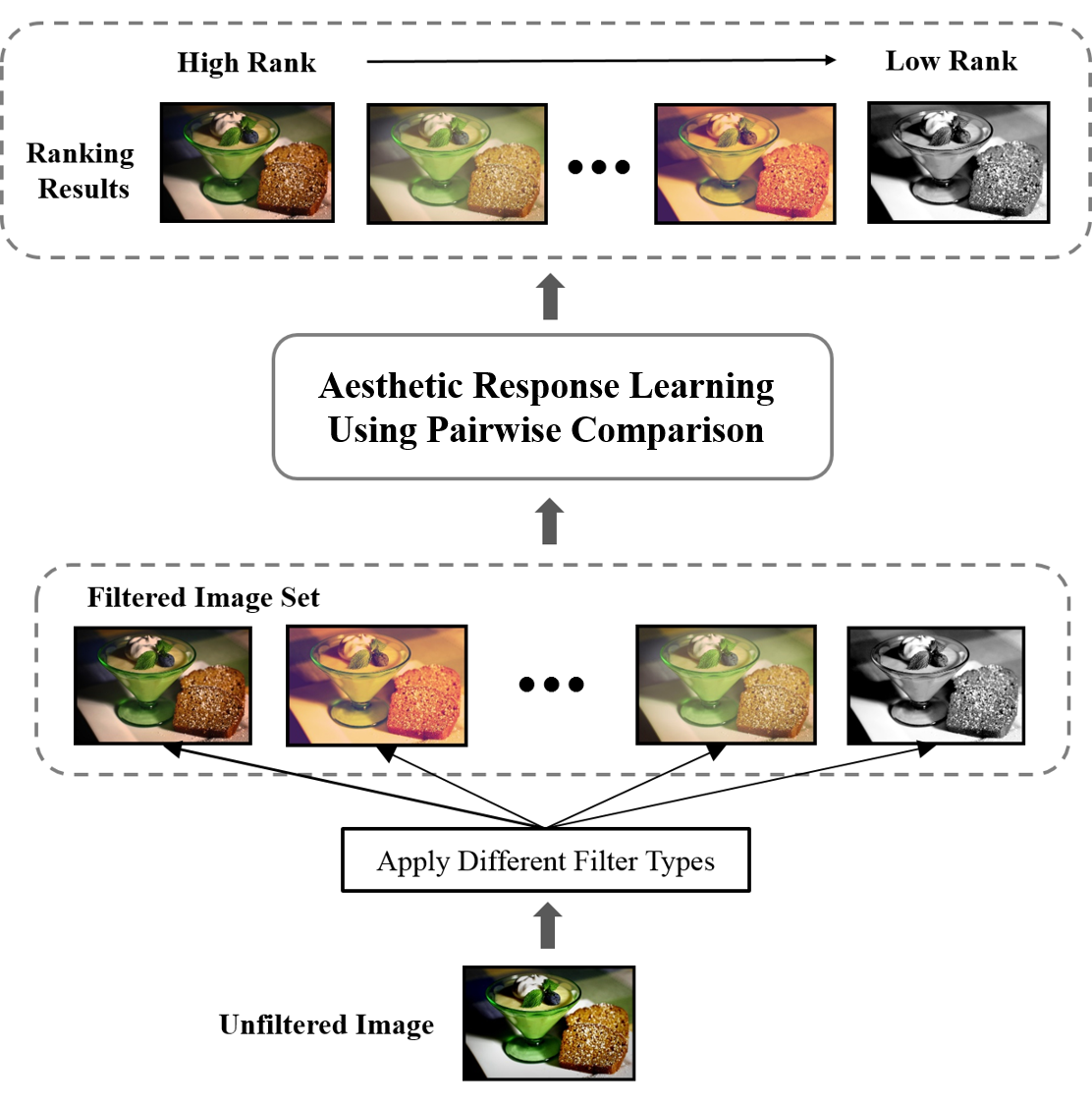}
\caption{Category-aware photo filter recommendation system. We propose category-aware aesthetic learning by utilizing our new collected pairwise labeled image dataset (FACD) for filter recommendation. Given an image, we generate a set of filtered images by applying different filter types. We then feed them into the system which learns image aesthetic using pairwise comparison (offline) and computes aesthetic responses for each filtered image. Finally, the system ranks filters by the aesthetic scores and the top-ranked filter is recommended.}
\end{figure}

\IEEEPARstart{W}{ith} the growth of social media population, people share and upload millions of photos per day.\footnote{https://www.instagram.com/press/} In addition to photo sharing, social media also provides photo filtering tools (e.g., Instagram, Flickr, Facebook) for users to enhance contrast, increase saturation, or change color tone for their photos \cite{bakhshi2015we, isola2011makes}. With the predefined filters (visual effects), users can spend less effort on stylizing or enhancing photos and achieve professional quality without image processing knowledge. Meanwhile, as reported in \cite{bakhshi2015we}, filtered photos will have higher chances of receiving views (+21\%) and comments (+45\%). Statistically, more than half of uploaded photos are filtered on Instagram\footnote{http://www.animhut.com/freebies/infographic/love-instagram-filter/}, a popular social media known for its image sharing functionality. Hence, manipulating images with filters becomes an essential function for photo sharing applications. In order to attract more users, social media has developed several types of filters. For example, Instagram provides more than twenty filters and frequently updates new one. Originally, the system is designed to allow users to select the best filter with a few clicks. However, they do not realize that users select favorite filters by comparisons. 

Due to the size restriction of portable device's display unit, users can only compare less than five filters at a time. They have to iteratively discard worst one until remaining filters can be compared at the same time. With the increasing number of filters, filter selection becomes a complex and time-consuming task. Hence, the need for efficient selection or recommendation of image filters is emerging. In December 2014, Instagram allows users to manage the order of filters based on their preference. However, we observe that users choose different filters based on image content (Sec. IV-C). It becomes a challenging problem to dynamically recommend image filters for users. To tackle the challenge, we aim to provide a filter recommendation system as shown in Fig. 1. The filtered images are sorted by their aesthetic scores learned from our collected pairwise labeled images with aesthetic judgment.

In quality learning field, Image Quality Assessment (IQA) has been studied for several years \cite{nriqa12, cbqa}. IQA focuses on classifying photos into two groups, high quality and low quality groups. The traditional approaches of IQA usually rely on the knowledge of photography (e.g., rule of thirds) and handcrafted features (e.g., color or SIFT) \cite{datta06, luo08, ke06}. More recently, machine learning is gradually becoming the main method for feature extraction and replacing handcrafted features. In \cite{dong2015photo, cnnnriqa}, deep learning is further introduced to the image quality problem. Photos are fed into Convolutional Neural Network (CNN) models to learn features and classify images. In spite of the general CNN structure, the learned features embed aesthetic information and improve classification accuracy.

Previous studies treat IQA as a regression problem that predicts images into different quality scores and further separates them into two classes. However, there is no clear boundary between high quality and low quality photos. It is a difficult task to distinguish images near the border even for humans. Therefore, an innovative perspective is proposed in \cite{quality14}. Chen \textsl{et al.} mention that image quality is the preference among images rather than an absolute aesthetic value of an image. That is, image aesthetic should not be quantified and mapped to the quality scores. The quality of an image is based on the comparison with another one. Thus, the regression problem can be transferred to the ranking problem which is solved by learning a pairwise ranking model \cite{yan2014learning}. However, these previous studies only use handcrafted features and conduct feature extraction and model learning separately. By using CNN, we can learn the model and features simultaneously. In addition, it is much simpler to select the better image between a pair than to pick out the best one among a pack of photos for humans.

Motivated by \cite{quality14, yan2014learning}, we propose our first-ever filter recommendation system based on pairwise aesthetic comparison in this paper. However, there is no existing and suitable dataset for this work. For pairwise aesthetic comparison with filtered images, we create a new dataset, Filter Aesthetic Comparison Dataset (FACD), generated from Aesthetic Visual Analysis (AVA) dataset \cite{ava12} which is widely used for image quality learning. The created FACD contains images with various filters and reliable image pairs with aesthetic judgment. Hence, we utilize the collected dataset and devise a novel pairwise aesthetic learning method for filtered images.

As mentioned in \cite{cbqa}, the professional photographers apply various techniques on different subjects. We have similar observations on the user preference of filtered photos for different image categories (e.g., flora, portrait) based on our pilot study. Hence, we further propose to integrate category (style) information with pairwise aesthetic comparison for multi-task learning. Therefore, for each image, we can obtain aesthetic responses and category information from the learned category-aware aesthetic model. For photo filter recommendation, we calculate aesthetic scores for different filtered photos and generate a filter aesthetic ranking for each unfiltered photo as shown in Fig. 1. The experiment results show that our method improves the recommendation on image aesthetic and outperforms other traditional methods designed for image quality prediction.

To sum up, the primary contributions in this paper include: 
\begin{enumerate}
\item Introducing a pairwise comparison method for filter recommendation based on convolutional neural networks. 
\item Utilizing the automatically predicted image category to improve the performance of filter ranking. 
\item Creating a new dataset, Filter Aesthetic Comparison Dataset (FACD), for pairwise aesthetic ranking. It contains 28,160 filtered images and 42,240 image pairs with user preference.\footnote{Available at http://wtwilsonsun.github.io/FACD/}
\end{enumerate}

In the next section, the related works of this paper are remarked. Then, we will introduce the collection of our dataset, FACD. After that, the network structure and methods are described in Sec. IV. Finally, we demonstrate experiment results and conclude the proposed method in Sec. V and VI.

%-------------------------------------------------------------------------
\section{Related Work}

\begin{figure*}[t]
\centering
\includegraphics[width=7in, keepaspectratio=true]{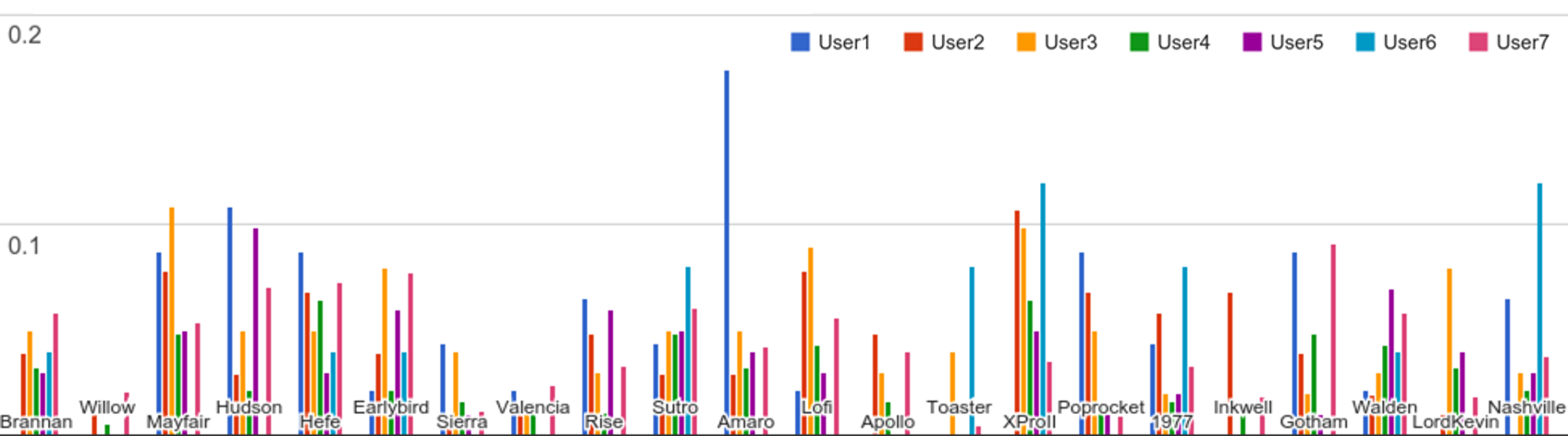}
\caption{User preference in pilot study. Each color depicts the preferred ratio of each filter to a specific user. It shows that user preference is diverse even though users have similar backgrounds. }
\end{figure*}

To provide photo filter recommendation with aesthetic learning, we first introduce the traditional image quality assessment (IQA). IQA can be separated into two parts in the past. One attempts to design algorithms based on photographic knowledge. For example, rule of thirds \cite{datta06, luo08} and simplicity \cite{ke06} are well-known techniques for image composition. Both of them concentrate on the subject in a photo. However, it is impossible to list all photographic skills exhaustively and implement them. Therefore, the other method for IQA is based on more general handcrafted features. Basic features including color \cite{datta06, ke06} and edge \cite{ke06} are commonly used. As the success of generic features in object detection and image classification, Marchesotti \textsl{et al.} \cite{marchesotti11} use SIFT, Bag-of-Visual-Words (BoVW), and Fish Vector for image quality classification. The handcrafted features are also used for view recommendation in \cite{su2012preference}. A manually designed feature, nature scene statistics (NSS), is also used by Mittal \textsl{et al.} \cite{nriqa12}.

Moreover, Convolutional Neural Network (CNN) has been applied on IQA problem recently. AlexNet \cite{imagenet12}, a breakthrough in computer vision using CNN, greatly outperformed the state-of-the-art methods which used handcrafted features before 2012. In these years, many variations of network structures have been presented, such as GoogLeNet \cite{googlenet15}, VGG \cite{vgg14} and Network-In-Network \cite{lin2013network}. These models achieve great performance in ILSVRC, which focuses on image classification, object recognition and localization. However, the purpose of image aesthetic learning is very different from object detection. The difference between them leads to the performance gap for a model applied on distinct domains. To deal with this problem, specific models should be designed for aesthetic learning. In \cite{rapid14, lu2015rating}, Lu \textsl{et al.} also notice that general architectures are not suitable for image quality classification which depends on both local and global information. They conduct experiments on network architectures by adjusting the number of layers. Eventually, they construct a CNN RAPID net \cite{rapid14} with four convolutional layers followed by three fully-connected layers for aesthetic learning specially. This structure is leveraged for advanced methods in this paper.

In \cite{quality14}, Chen \textsl{et al.} propose a ranking strategy for image quality assessment. They first extract handcrafted features for input images, and then train a rankSVM to learn a ranking function for image comparison. However, this method is separated into two stages, feature extraction and model learning. Inspired by \cite{quality14}, we extend the traditional aesthetic classification issue to the pairwise comparison problem using CNN which extracts features and learns the model simultaneously. General studies \cite{cnnnriqa, dong2015photo} focus on single-column quality classification, but the double-column network is needed for pairwise input. In \cite{chopra2005learning}, Siamese Network, which is widely used for similarity learning and face retrieval, provides an example of a double-column network. It embeds metric learning in CNN by contrastive loss, a distance function for relevant and irrelevant pairs. Therefore, the activations of hidden layers are used as representations for recognition and retrieval. Further, \cite{deeprank14, hoffer2015deep} even extend the double-column architecture to the triplet network which learns features from both positive and negative samples. However, the contrastive loss and triplet loss do not fit the objective of pairwise aesthetic comparison in this work. To learn preference from image pairs, we formulate a novel loss function that compares aesthetic responses and construct the double-column model for pairwise input.

In addition, multi-task learning is also adopted in object detection and facial landmark detection in \cite{girshick2015fast, ren2015faster, zhang2014facial}. The output of last fully-connected layer is regarded as the representation and is directed to multiple loss layers with distinct objective functions. The performance of main task can be further improved by the related minor tasks. This strategy also avoids the complex design and long training period of sequential training process as all tasks are learned at the same time. In this work, we assume that the filter preference is related to image category (content). As a result, we introduce multi-task learning to our method for aesthetic learning and category classification.

%-------------------------------------------------------------------------
\section{Filter Aesthetic Comparison Dataset (FACD)}
Since there is no dataset designed for filter aesthetic ranking, we create our own dataset and evaluate the proposed method on it. In this section, we first describe the generation of filtered images. Then the pilot study for filter preference investigation and the online crowdsourcing for pairwise filter annotations are introduced.

\subsection{Filtered Image Collection}
First, we collect a set of unfiltered images, also called reference images in this paper, from an existing dataset. The reference images are obtained from AVA dataset \cite{ava12}, a large-scale dataset for visual aesthetic analysis, which contains over 250,000 photos for aesthetic study. The photos are divided into more than 60 categories with semantic labels in the AVA dataset. We sample our reference images from the top 8 most popular categories which are equivalent to the ones used in \cite{cbqa, ava12}. The categories include \textit{animal, flora, landscape, architecture, food and drink, portrait, cityscape, and still life}. For each category, we randomly pick 160 photos. Therefore, we totally collect 1,280 unfiltered images in 8 categories from the AVA dataset. Next, we define the filter types for the production of filtered images. Because the filter types on social media are time-varying, we choose 22 filters\footnote{1977, Amaro, Apollp, Brannan, Earlybird, Gotham, Hefe, Hudson, Inkwell, Lofi, LordKevin, Mayfair, Nashville, Poprocket, Rise, Sierra, Sutro, Toaster, Valencia, Walden, Willow, and XProII.} provided by both GNU Image Manipulation Program (GIMP) toolkit\footnote{https://www.gimp.org/} and Instagram to simulate the real situation of filter selection on social media. All of the filters are applied on each reference image and then the dataset has 28,160 filtered images in total.

\subsection{Pilot Study on Filter Preference Investigation}
With these filtered images, we first investigate the user preference on filtered images. Hence, we conduct a pilot study on a small group of participants. All of them have similar backgrounds: master students, aged 22-26, and also Instagram users. Since it is complicated and time-consuming for users to select the best filtered images from 22 filters, our pilot study is designed as a pairwise comparison problem motivated by \cite{quality14}. For this study, each of them is given a pair of filtered images (e.g., the top two images in Fig. 3) and is asked to select the better image each time. Fig. 2 shows the distribution of filter preference on distinct participants. Each user is depicted in one color and the histogram represents the selected ratio of each filter. The figure demonstrates that filter preference among the participants is diverse in spite of their similar backgrounds. That is, the filter selection is greatly subjective and would not be influenced by the user background. Therefore, we can choose anyone to generate the comparison label without considering his/her background or experience. Based on the pilot study, it motivates us to conduct a comprehensive analysis of user preference on different filters and image categories. Hence, we attempt to collect a large number of filtered image pairs with aesthetic judgment. Thus, we design a more rigid aesthetic judgment and annotation in the next section.

\begin{figure}[t]
\centering
\includegraphics[width=3.7in, keepaspectratio=true]{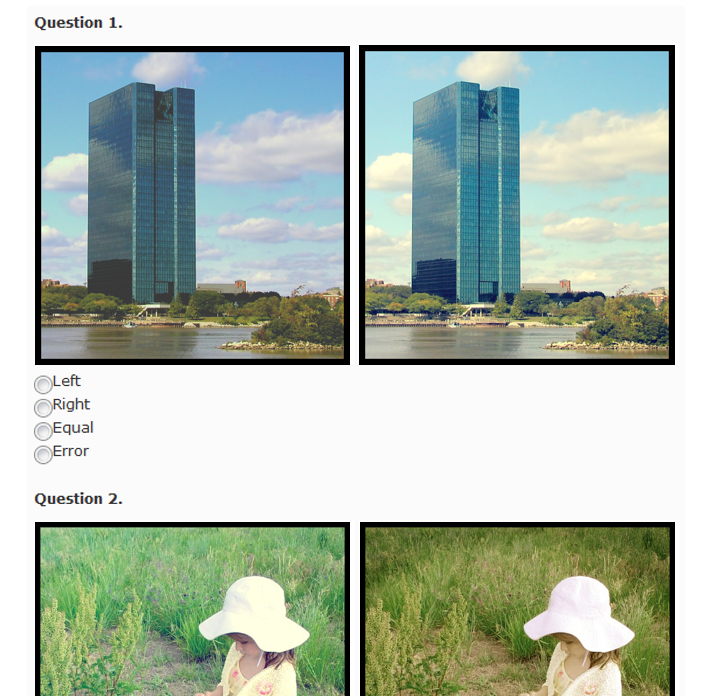}
\caption{The interface for crowdsourcing annotation on Amazon Mechanical Turk. Each question provides a pair of filtered images and four options. Participants are asked to vote for the preferred photo based on their aesthetic judgment.}
\end{figure}

\subsection{Pairwise Image Comparison on Amazon Mechanical Turk}
To effectively obtain enormous amount of filter preference, we establish aesthetic annotation on filtered images via crowdsourcing approaches. In \cite{crowdsourcing14}, Redi \textsl{et al.} study the performance of both paid subjects and volunteers for image aesthetic rating. The results show that the volunteer crowd is more reliable but leaves more incomplete tasks than the paid one. Compared with the rating problem in \cite{crowdsourcing14}, pairwise quality ranking is a simpler task so we assume that the manual annotation only contains a small portion of noisy or incorrect labels. Meanwhile, it is also infeasible to obtain complete order of filter preference for each reference image because it is time-consuming to decide a suitable ranking. Hence, we approximate the ideal ranking order based on the following criteria. First, the number of occurrence for each filtered image should be balanced in the sampled pairs. Meanwhile, all desired image pairs should be labeled as complete as possible. Therefore, we appeal to paid participants for image aesthetic comparison on Amazon Mechanical Turk (AMT) \cite{mason2012conducting}. 

For each reference image, it has 22 corresponding filtered images so there are ${22 \choose 2}$ combinations at most. However, annotating all pairs on AMT is still an expensive and time-consuming task. Thus, we only generate 33 pairs for a reference image from its corresponding filtered image set and make sure that each filtered image appears in three pairs.\footnote{Note that we assume filter preference is transitive. That is to say, if we know that $\mathit{filter}_A > \mathit{filter}_B$ and $\mathit{filter}_B > \mathit{filter}_C$, we will assume $\mathit{filter}_A > \mathit{filter}_C$. Hence, we can reduce the annotation number of filter pairs and utilize the preference ranking scores (Sec. V-A) to decide ground truth filters for each image in our experiments.} The annotation task is demonstrated in Fig. 3. Our designed task which is also called a ``HIT" (Human Intelligence Task) contains 10 questions of pairwise comparison. Each question consists of two filtered images with four options, including left, right, equal, and error for image display problem. Users need to select one option as the answer for a question according to his/her aesthetic reception. For each pair, the selected (preferred) image is viewed as a positive image, whereas unselected one is viewed as a negative image. 

\begin{figure*}
\centering
\includegraphics[width=7in, keepaspectratio=true]{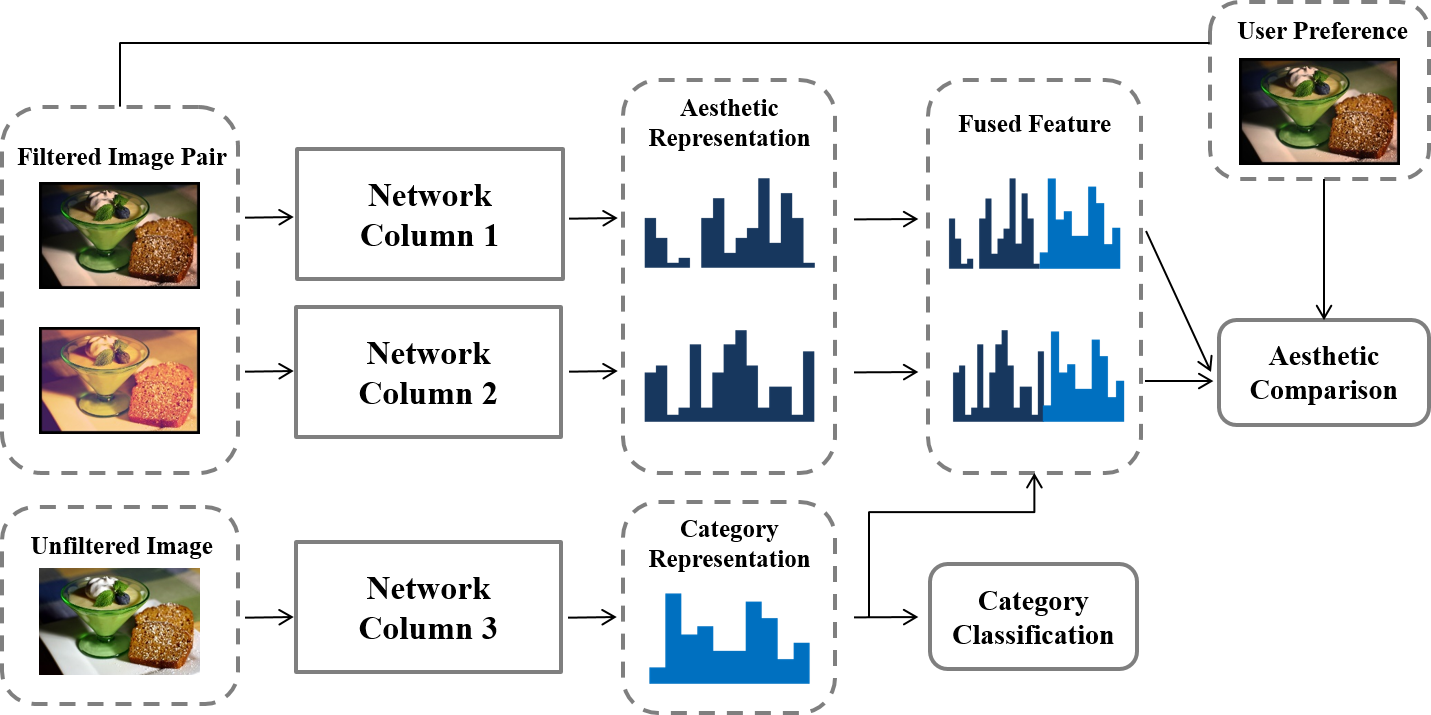}
\caption{System overview of the proposed category-aware aesthetic representation learning on filtered images. The training model contains aesthetic comparison and category classification. The top two networks are designed for learning the aesthetic responses and quality comparison by leveraging user preference on filtered images (aesthetic comparison). The bottom one is used for category classification. As discussed in Sec. IV-C, the category information (e.g., food, landscape) will affect the user preference on selecting proper filters. Hence, we integrate the category representation into the aesthetic responses to learn category-aware aesthetic representations. Note that the three network columns share parameters so that the order of images in a pair is arbitrary. }
\end{figure*}

To ensure the credibility of the user annotation, we adopt three methods to avoid unreliable labels: 1) check answer completeness, 2) check label consistency, and 3) check verification. For the first method, we guarantee that all questions in a HIT are completely answered since all filtered images are fairly sampled for three times in pairs. Moreover, ambiguous answers are meaningless to the comparison problem in our dataset. Thus, a task would be rejected if a worker selects more than one ``Equal" option in a single HIT. Second, we place a duplicated question in each HIT to filter out the users with inconsistent decisions. The duplicated question is the same as one of the other nine pairs in a HIT, except we interchange the order of the two filtered images (e.g., reverse the top two images in Fig. 3). To maintain the annotation consistency, users with distinct labels between the original pair and the duplicated one are regarded as doubtful participants, and we will ask other users redo these tasks. Last, we provide a mathematical question in the end of a HIT to avoid robotic answers. A HIT without correct mathematical answer is also rejected. Similarly, new users will annotate rejected tasks again until all pairs are annotated. 

Finally, the whole FACD contains 1,280 reference images collected from eight categories, 28,160 filtered images created from 22 filters, and 42,240 filtered image pairs with aesthetic comparison labels. It also contains the comparison scores and classes (qualities) for each filtered image. The generation of scores and classes will be described in Sec. V-A. 

%-------------------------------------------------------------------------
\section{Category-Aware Aesthetic Representation Learning}
To analyze user preference on filtered images, we propose category-aware aesthetic representation learning by utilizing our collected FACD. Fig. 4 shows the system overview of the proposed method. We attempt to learn aesthetic responses between a pair of images using CNN models that embeds aesthetic information in the hidden layers. Since the structure of CNN varies from task to task, a proper structure should be selected for aesthetic learning in this work. Furthermore, the objectives of typical loss functions are different from the target of pairwise comparison. Hence, we formulate a customized loss function to deal with the comparison problem. To improve the result of filter recommendation, we further combine the basic double-column model with an image category classification task. In this section, we first describe the network structure we use in this paper. Next, the concept and formulation of the loss function are shown in detail. Finally, the multi-task learning integrating pairwise comparison and category classification will be introduced in the end.

\subsection{CNN Structures}
As reported in recent works \cite{imagenet12, googlenet15, rapid14}, deep learning is vary promising in various research areas because it can learn feature representations and model parameters simultaneously. Hence, we devise our learning method based on deep learning, and investigate the effect on different CNN structures. AlexNet \cite{imagenet12} has been widely referenced in different domains. Based on the large-scale image classification dataset, AlexNet is frequently used for fine-tuning and regarded as the simplest design for CNN-related works. Thus we also take AlexNet as a reference model for the baseline. However, it is designed for object detection and classification originally. There is no specific structural design for aesthetic in AlexNet. In order to compare with AlexNet, the non-specific model for aesthetic learning, we also introduce another CNN structure that is designed for image quality classification. 

\begin{table}
\centering
\caption{Network details. The networks are constructed based on AlexNet\cite{imagenet12} (left) and RAPID net \cite{rapid14} (right). The difference is that the last fully-connected layer for classification is removed and the dimension of the last fully-connected layer is reduced to 128 which represents the aesthetic responses of an image. }
\begin{tabular}{cc|cc} \hline
Layer & Kernel Size & Layer & Kernel Size\\ \hline
conv1 & 11$\times$11$\times$96 & conv1 & 11$\times$11$\times$64 \\ \hline
pool1 & 3$\times$3 & pool1 & 3$\times$3\\ \hline
conv2 & 5$\times$5$\times$256 & conv2 & 5$\times$5$\times$64 \\ \hline
pool2 & 3$\times$3 & pool2 & 3$\times$3 \\ \hline
conv3 & 3$\times$3$\times$384 & conv3 & 3$\times$3$\times$64 \\ \hline
conv4 & 3$\times$3$\times$384 & conv4 & 3$\times$3$\times$64 \\ \hline
conv5 & 3$\times$3$\times$256 & - & - \\ \hline
pool5 & 3$\times$3 & - & - \\ \hline
fc1 & 4096 & fc1 & 128 \\ \hline
fc2 & 128 & - & - \\ \hline\end{tabular}
\end{table}

In \cite{rapid14}, Lu \textsl{et al.} conduct several experiments on CNN structures for aesthetic categorization. They experiment on different combinations of layers and then propose an aesthetic-oriented model. Their proposed network, named RAPID net in this paper, is composed of four convolutional layers and three fully-connected layers. Particularly, they apply a pooling layer and a normalization layer on the first two convolutional layers. In comparison with AlexNet, the depth is shallower and the number of kernel is fewer in RAPID net. Since image quality considers more details and local regions than the global view, the shallower model learns more detailed information for aesthetic. The fully-connected layers in RAPID net also have fewer neurons than AlexNet. Each dimension in fully-connected layer represents certain aesthetic responses, whereas it learns object patterns in AlexNet. Thus, the feature dimension is reduced from 4,096 to 256 or 128 for aesthetic learning.

To compare the performance of distinct networks, we take both AlexNet and RAPID net as reference models and adjust the details to fit our purpose. The network details are shown in Table I. In our model, we remove the last fully-connected layer which is designed for image classification, and reduce the output dimension of the last layer to 128 for learning more specific aesthetic responses. In addition, we only retain two and one fully-connected layers for AlexNet and RAPID net respectively. The reduction of fully-connected layer leads to less parameters and avoids overfitting as well.

\begin{figure*}
\centering
\includegraphics[width=6.7in, keepaspectratio=true]{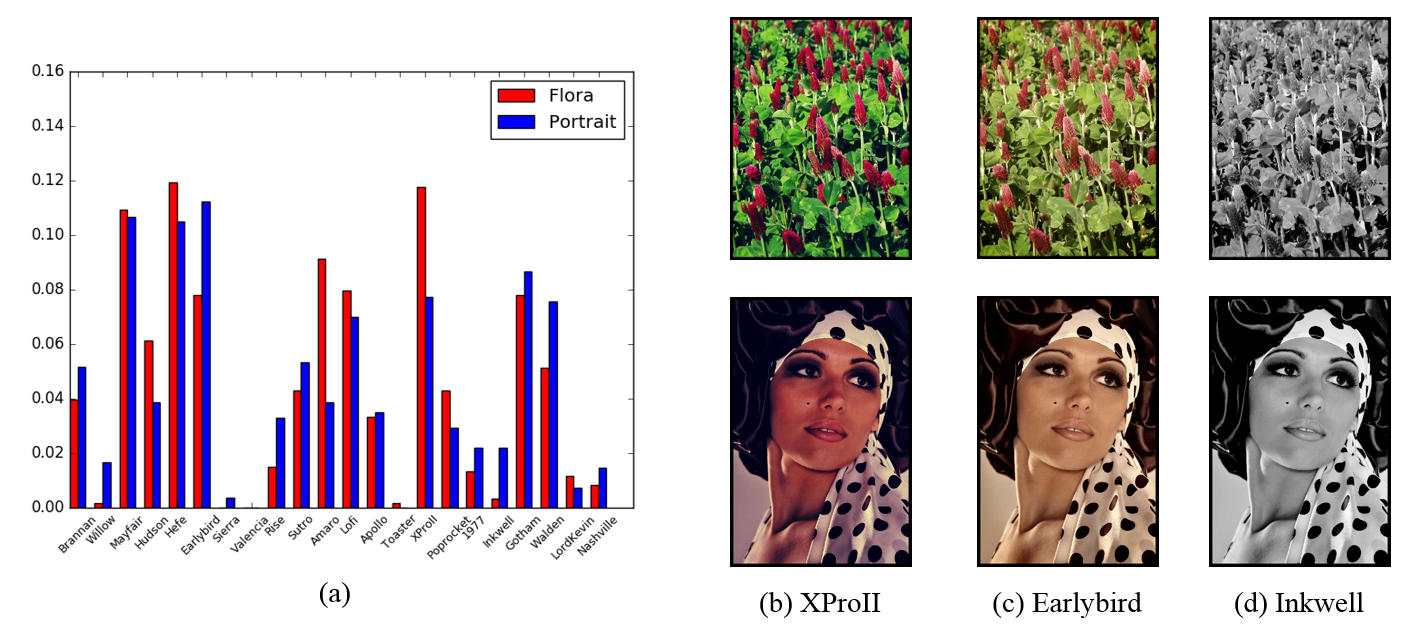}
\caption{(a) The distribution of filter preference in different categories. The chart illustrates the selected ratio of each filter in flora and portrait categories. It shows that the filter selection is sensitive to the image category. The examples on the right are images of flora and portrait, respectively. We apply three filter types (b) XProII, (c) Earlybird, and (d) Inkwell. The examples reveal that the image attractiveness varies with image category even if the same filter is used. }
\end{figure*}

\subsection{Aesthetic Response Learning by Pairwise Comparison}
Inspired by \cite{quality14}, we take aesthetic learning as a pairwise ranking problem between images rather than classic quality classification. To support the input of image pairs, we extend the single-column network to double-column one like the architecture in \cite{rapid14}. Each of the networks in Table I is duplicated and combined with the reproduction in parallel (i.e., Network Column 1 and 2 in Fig. 4). The parameters of these network columns are shared between the same layers as typical multi-column approaches. Since CNN model can embed aesthetic information in the learned representation automatically, the representation can be regarded as the vector of aesthetic responses. Hence, images with higher aesthetic responses indicate they are more visually appealing. We can rank images by the intensity of this representation. To achieve this objective, we propose aesthetic response learning by pairwise comparison (PairComp), and attempt to maximize the difference in aesthetic responses ($||f||^2$) between filtered image pairs. The proposed formulation is defined as
\begin{equation} \label{eq:paircomp}
\textcolor{black}{Max \sum_{i=1}^N D(s_i) = Max \sum_{i=1}^N (||f_{i,p}||^2-||f_{i,n}||^2),}
\end{equation}
where $f_{i,p}$ and $f_{i,n}$ denote the aesthetic representations of positive (p) and negative (n) images in the \textit{i}-th sample pair ($s_{i}$) respectively. As shown in Fig. 4, based on the given user preference, the above formulation learns to maximize responses of positive images $f_{i,p}$ and minimize the intensity of negative images $f_{i,n}$ simultaneously. As the positive images have greater aesthetic responses, the difference would be larger. To integrate with general CNN frameworks, we can use negative $D(s_{i})$ as the loss function (i.e., $Min \sum_{i=1}^N -D(s_i)$). By minimizing the loss, the model learns to produce larger aesthetic responses for the better photo (preferred filter). Furthermore, rather than calculating the difference in learned feature representations ($f$) between filtered images (i.e., $||f_{i,p}-f_{i,n}||^2$), we propose to measure their aesthetic responses (i.e., $||f_{i,p}||^2-||f_{i,n}||^2$). Hence, it enforces those preferred filters will have higher aesthetic scores, and we can directly utilize the learned aesthetic scores ($||f||^2$) to rank and decide the user preference without pairwise comparison.

\subsection{Category-Aware Filter Preference Learning}
In addition to the pairwise preference of filtered images in our collected FACD, the reference images collected from AVA dataset also consist of the category information for each image. We observe that not only filter types (e.g., XProII, Amaro) but also image categories (e.g., flora, landscape) affect user preference on filtered images. To illustrate the influence of image category, we show the distribution of filter preference for portrait and flora categories in Fig. 5 (a). The chart confirms the assumption that the preference of filters depends on the image category. In Fig. 5 (b)-(d), we also show some filtered results on flora (the first row) and portrait (the bottom row) images. For flora, the focus is always on the colorful subjects. If the visual effects of filters are designed to enhance contract or lighting (e.g., XProII), those flora images can be highlighted (preferred by users) as the example in Fig. 5 (b). On the other hand, users prefer portrait photos with filters that offer more warm color for skin tone in the photo (e.g., Earlybird). Compared with Fig. 5 (c), Fig. 5 (b) making the skin fuscous is less attractive to users. Though colorful images are more likely preferred, some colorless photos are attractive due to special effects. For instance, Inkwell filter (Fig. 5 (d)), which converts the colorful image into black and white, supplies vintage view to portrait photos but loses the main focus in flora images.

To take image category into account for filter recommendation, we utilize this external information for multi-task learning. In addition to the double-column model, we construct an additional network for image categorization. The complete network design that contains aesthetic learning and category classification is shown in Fig. 4. As mentioned in the previous section, the three network columns have the same structure and share all parameters. The top two networks are fed with image pairs and learn pairwise aesthetic responses. The remaining network (Network Column 3 in Fig. 4) for category learning takes the reference (unfiltered) image from each filtered pair and the corresponding category label as input in training. After the input images are forwarded to the networks, the aesthetic representation is concatenated with the category representation. We further apply a fully-connected layer on the concatenated features to generate fused features. Hence, the fused features containing both aesthetic and category information are used for aesthetic comparison. In the last layer of aesthetic comparison, the pairwise loss is calculated using the labels gathered on AMT as described in Sec. IV-B. Hence, the loss function for category-aware filter preference learning (PairComp+Cate) is defined as
\begin{equation}
\textcolor{black}{Min \sum_{i=1}^N [-D(s_i) + SoftmaxLoss(u_i)],}
\end{equation}
where SoftmaxLoss() measures the classification error for the given unfiltered image ($u_i$) and the first term calculates the difference in aesthetic responses between filters as mentioned in Eq. (\ref{eq:paircomp}). In backpropagation, the gradient of the fused features is passed to the previous layers. The weights are updated for both aesthetic learning task and category classification task.\footnote{In this work, we are to investigate the effect of aesthetic responses under CNN structures so we mainly focus on utilizing additional information (i.e., category information and pairwise aesthetic comparison) for better filter recommendation. Therefore, we do not integrate our proposed method with other possible loss functions (e.g., \cite{Weston10}) for aesthetic learning.} It makes the model revise parameters and improve the performance of filter ranking. The improvement of utilizing category prediction will be demonstrated in Sec. V-B.

%-------------------------------------------------------------------------
\section{Experiments}
We evaluate the proposed method on our dataset, FACD. In this section, we first describe how we generate the ground truth from FACD and leverage the dataset for both training and testing. The process design for both training and testing phases is also introduced. Next, we show the experiment results of the proposed methods. The results of different network structures and distinct methods are compared in detail. Then, we illustrate examples and discuss our observation. 

\subsection{Setting}
\subsubsection{Dataset}
In our created Filter Aesthetic Comparison Dataset (FACD, Sec. III), there are totally 42,240 image pairs with human labels generated from 1,280 reference images. We first separate the dataset into two parts for training and testing (7:1). The 1,280 reference images are divided into 1,120 and 160. The 1,120 unfiltered images and their corresponding filtered image pairs comprise the training set. That is, there are 36,960 image pairs used as training data. The remaining images and pairs are used for testing and ground truth generation. To evaluate the performance of filter recommendation on the created FACD, we attempt to generate a ranking list for each unfiltered (reference) image. Since there are only pairwise comparison labels in FACD, we have to define the ground truth of ranking results based on these pairwise labels. In Sec. III-C, we ensure that each filtered image appears in three comparison pairs. For each pair, we give a positive score (+1) for the preferred (selected) image, otherwise a negative score (-1). Note that if users select `equal' in the annotation process, both images will be assigned zero score. Hence, each image receives a score which represents its quality ranging from -3 to +3 since it appears in different pairs for three times. We take the filtered images rated +3 (i.e., beat all compared filters) as ground truth for each reference image. As there may be more than one image with +3 scores, the average number of ground truth is more than one. In our dataset, each reference image has 3.7 filtered images as ground truth in average.\footnote{As demonstrated in our pilot study in Sec. III-B and Fig. 2, we find that user preference on image filters is diverse. Different users would prefer different types of filters for the same images. Therefore, for each image, the preferred (ideal) filters might contain multiple choices. Hence, the averaged 3.7 ground truth filters would still be a reasonable number for evaluation.}

\subsubsection{Training}
For aesthetic comparison, image pairs are directly fed into the double-column CNN model with labels (user preference) in the training phase. The input images are resized to 256$\times$256 in advance. In each iteration, images of the input batch are randomly cropped into 227$\times$227 and 224$\times$224 for AlexNet and RAPID net separately according to the original design in \cite{imagenet12, rapid14}. Data augmentation is also adopted by reflecting images horizontally. To further improve the ranking result as proposed in Sec. IV-C, the corresponding reference image of the training pair is also fed into the network for category classification at the same time. We conduct experiments on Caffe \cite{jia2014caffe}, a widely used framework for deep learning, for both training and testing. 

\begin{table}[t]
\centering
\caption{Testing accuracy of filter recommendation. ``PairComp" denotes the implementation of pairwise comparison loss (Sec. IV-B) and ``+Cate" means training with category classification (Sec. IV-C). The last two rows indicate that our method outperforms the traditional CNN model (AlexNet) and the aesthetic-oriented model (RAPID net). We can achieve better results than the model designed for object detection. More details are described in Sec. V-B. }
\begin{tabular}{c|c|c|c} \hline
Model & Top-1 & Top-3 & Top-5\\ \hline
Random Guess & 16.80\% & - & - \\ \hline
AlexNet\cite{imagenet12} & 33.13\% & 70.63\% & 88.75\% \\ \hline
RAPID net\cite{rapid14} & 37.50\% & 72.50\% & 86.25\% \\ \hline\hline
PairComp+SPP2 & 36.88\% & 66.88\% & 79.38\% \\ (AlexNet) & & & \\\hline
PairComp+SPP3 & 38.13\% & 69.38\% & 83.13\% \\ (AlexNet) & & & \\\hline
PairComp & 38.75\% & 78.13\% & 88.75\% \\ 
(AlexNet) & & & \\ \hline
PairComp & 41.25\% & 78.13\% & 88.13\% \\ 
(RAPID net) & & & \\ \hline\hline
PairComp+Cate & 41.25\% & {\bfseries 80.00}\% & 89.18\% \\
(AlexNet) & & & \\ \hline
PairComp+Cate & {\bfseries 41.88\%} & 79.50\% & {\bfseries 90.00\%} \\ 
(RAPID net) & & & \\ \hline\end{tabular}
\end{table}

\subsubsection{Testing}
The aesthetic representation is extracted from the last fully-connected layer (i.e., 128-d in our experiments) for filter recommendation. As all the networks are shared parameters, we only utilize the model (single network) to extract aesthetic representations for all filtered images (e.g., 22 filters). Meanwhile, it is unnecessary to go through the comparison loss layer in testing phase, and the input only takes a single filtered image rather than a pair. Even for the multi-task model, only an unfiltered image is needed additionally for fused features. Then the ranking result is generated by sorting the Euclidean norm (i.e., aesthetic scores = $||f||^2$) of the representations. It is more efficient than traditional pairwise ranking which compares all combinations of pairs. Based on our defined ground truth in the previous section, we can evaluate the success rate (accuracy) of the filter recommendation system. Hence, if the ground truth filters appear in the top \textit{K} results, the ranking results are regarded as correct answers. In the experiments, we set K to be 1, 3, and 5 because users can only see less than five filters simultaneously on their smart devices. This setting is meaningful to the recommendation system since it guarantees that at least one correct filter appears in the first page of the filter selection interface.

\begin{figure*}[t]
\centering
\includegraphics[width=7in, keepaspectratio=true]{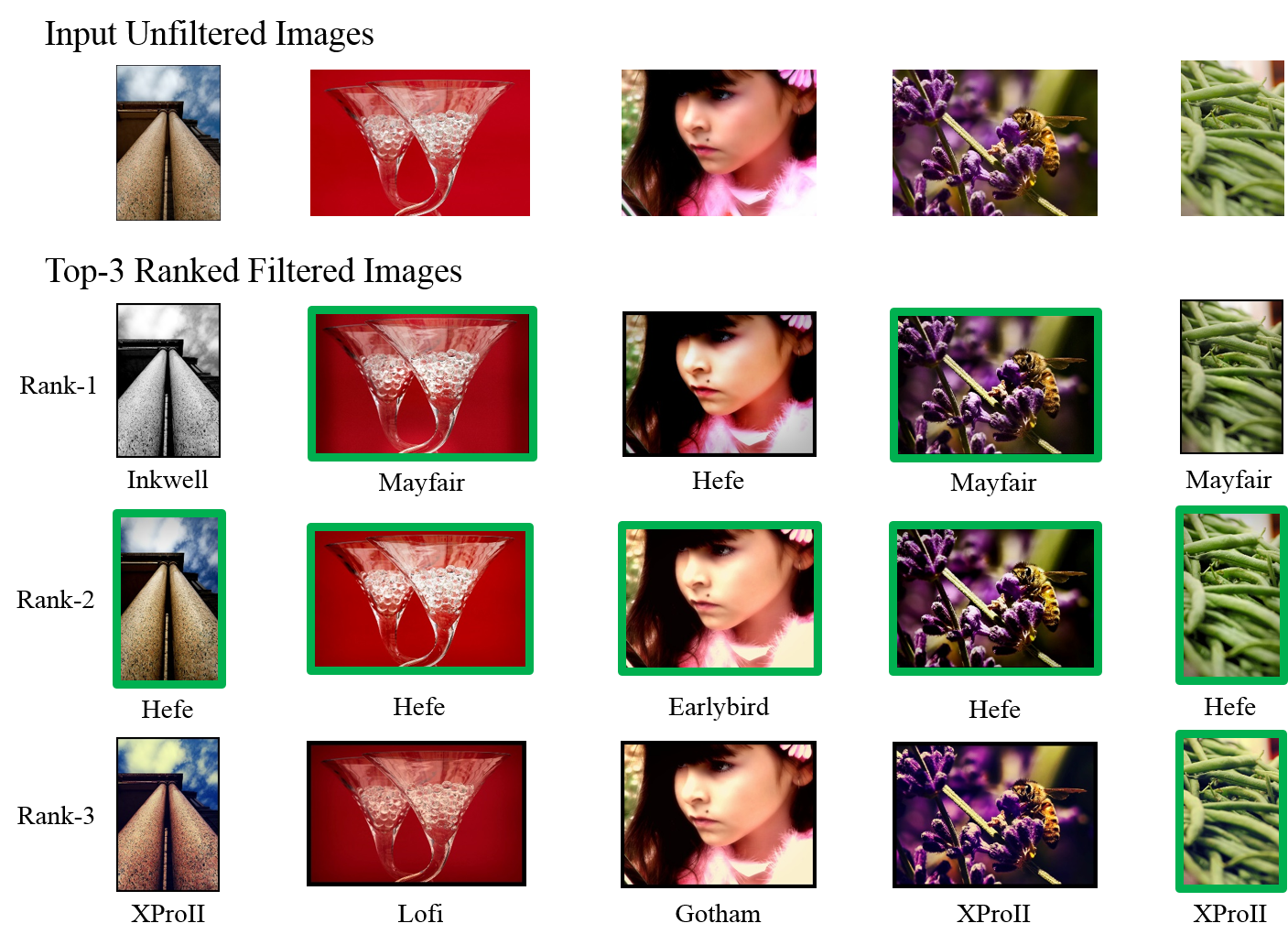}
\caption{Examples of top-3 recommended filters on the proposed category-aware aesthetic learning model (PairComp+Cate [RAPID net]). Each column presents an unfiltered image and its corresponding filter recommendation. The categories of the examples (from left to right) include (a) architecture, (b) still life, (c) portrait, (d) flora, and (e) food \& drink. The images with green bounding boxes are correct predictions (preferred/ideal filters). The figure shows that the filter preference depends on the image content and the characteristic of image category, and our model also learns some of the properties automatically. More explanation is discussed in Sec. V-C.}
\end{figure*}

\subsection{Results}
We propose two methods for aesthetic learning in Sec. IV-B and IV-C. To evaluate the improvement of each method, we utilize different network structures and compute top-1, top-3, and top-5 accuracy for each case. Table II shows the overall comparison on different structures. The models denoted as ``PairComp" use the pairwise comparison loss (Sec. IV-B) in the training phase and the ones named ``+Cate" are integrated with category classification (Sec. IV-C). 

\subsubsection{Baselines}
The first 3 rows of Table II are the results of baselines, including random guess and single-column CNNs. The result of random guess (16.8\%) comes from averaged 3.7 ground truth divided by 22 filter types. The basic CNNs are constructed from Table I and learn binary classification for image quality as in \cite{rapid14, lu2015rating, dong2015photo}. They only classify images into high quality or low quality directly. To generate the ranking prediction, we take the probability of high quality as the aesthetic score of a filtered image. Then the ranking is obtained by sorting the scores. It is obvious that the aesthetic-specific model (RAPID net) outperforms the typical CNN architecture (AlexNet) designed for object classification. It means that RAPID net can embed more aesthetic information in the representation. 

\subsubsection{Models using Pairwise Comparison (PairComp)}
The following four rows of Table II demonstrate the accuracy of pairwise models combined with the aesthetic comparison loss. As Lu \textsl{et al.} \cite{lu2015deep} suggested that the details of an image also impact the aesthetic classification, we also conduct our experiments on the original images (i.e., without resizing). To deal with varying image sizes in our dataset, we insert the spatial pyramid pooling (SPP) layer \cite{spp15} with different number of levels between the pooling layer and the fully-connected layer on AlexNet (i.e., between pool5 and fc1 in Table I). In Table II, the number after the term ``SPP'' means the levels of the SPP layer. However, the SPP layer is designed for image classification and may not be suitable for the pairwise comparison problem of image quality. It only outperforms the basic single-column model on top-1 accuracy but even lowers the performance in the other two cases. Though the SPP layer fails to improve the prediction for filter ranking, other results still shows that our proposed pairwise comparison loss layer supports aesthetic response learning. The accuracy of both PairComp (AlexNet) and PairComp (RAPID net) obtains about 4-5\% improvement. The results verify the proposed aesthetic-specific structure can learn features with more quality information than general CNN structures (e.g., AlexNet).

\subsubsection{Models using Pairwise Comparison and Category Classification (PairComp+Cate)}
At the bottom of Table II, the results of pairwise aesthetic ranking with category are presented. Even though the improvement of integrating image category is less than the pairwise comparison, it still increases the accuracy about 1-3\%. We believe that a category-specific structure for the third network column can classify the image category more accurate and then improve the ranking results. With both pairwise comparison and category classification, the top-1 accuracy of both AlexNet and RAPID net exceeds 40\%. Besides, the top-3 and top-5 accuracy are close to 80\% and 90\% respectively. It indicates that the recommendation system can provide a suitable filter for an input image to a user on the display with limited size (e.g., 5 filters).

Meanwhile, we observe that users usually prefer and focus on few filters although there are lots of different filter types. Based on the statistics and analysis from Marketo, it shows that only top few filters have higher usage (e.g., 10\% for Earlybird, 8\% for XProII, and 5\% for Valencia).\footnote{http://blog.marketo.com/2013/03/what-your-instagram-filter-says-about-you-infographic.html} Therefore, user preferred filters are prone to show up in the top-5 results, and the improvement is small. As the number of filters increases, we believe that the proposed category-aware aesthetic learning can achieve better accuracy in top-5 accuracy. If we can obtain the full rank of 22 filtered images, we can learn a better model for the filter recommendation. However, it is time-consuming to collect all possible ranking for the evaluation. In this work, we attempt to leverage a small number of labeled filter pairs for aesthetic learning.

\subsection{Observation and Discussion on Filters and Categories}
We also visualize some filter recommendation results on our proposed PairComp+Cate (RAPID net) in Fig. 6. It shows the testing (unfiltered) images along with their corresponding top-3 recommended filters. The images with green borders represent the correct predictions (ideal filters) of the given images. We find that filter preference of \emph{architecture} and \emph{still life} depends on the texture of subjects. For instance, materials like tile, glass and metal are more sensitive to light reflection on their surface. Besides, the examples support the observation mentioned in Sec. IV-C. The colorful images, such as flora photos in the fourth column of Fig. 6, become more elegant by enhancing contrast or lighting to emphasize the subjects. The flora images with Hefe filter show the improvement of stronger contrast. This principle can be applied on \textit{food and drink} as well. The more colorful food images seems tastier to humans. On the contrary, flora with XProII filter makes it difficult to distinguish the main object.

The images of portrait in the third column of Fig. 6 also illustrate their dependency on color tone and brightness. Because of the skin color of human being, a warmer tone (e.g., Earlybird filter) can yield more vitality to the images of portrait. By contrast, cool tone in these kinds of images brings about negative feeling to viewers. Besides, lighting is another factor impacting the preference especially for portrait photos. The impact of brightness reflects on skin appearance mainly. In the figure, the correct prediction (Earlybird) of portrait is brighter than others obviously and the skin seems shinier and smoother. In addition, we think the portrait image with Gotham filter is also visually appealing; however, the preference is still subjective and sensitive to the sampled pairs (i.e., better than 3 filtered images).

In addition to the image examples, we also compare the filter distribution between our prediction and the ground truth as shown in Fig. 7. The red bars depict the preference of filters in the ground truth. We find that some filter types are more attractive to users specifically, such as Hefe, Mayfair, and Gotham. With further survey of these filters, both of Hefe and Mayfair enhance the contrast of color and Gotham transfers images to warm color. This meets the explanation of Fig. 6 so that these filters are more generally applied on images. Besides, the blue bars illustrate the filter distribution of our prediction. Despite of the imbalance of user preference, the CNN model still learns aesthetic information by the proposed method. The distribution of predicted (recommended) filters is similar to the user preference in the ground truth, except Gotham. XProII also provides warm color and high contrast effects like Gotham. Therefore, images in Gotham may be incorrectly classified into XProII. From Fig. 7, we find that high contrast filters are easier to be selected correctly. It means that XProII is recommended because of its high contrast rather than warm color in most cases. It indicates that our model is more sensitive to contrast than to warm color. How to deal with the filters with great difference between ground truth and prediction is a direction for further study in the future.

For comparing with traditional approaches, in \cite{rapid14}, Lu~\textsl{et~al.} compare the performance of handcrafted features, such as Fisher Vector (FV), and the CNN models on the AVA dataset.\footnote{Besides the non-CNN approach on AVA dataset, Guo and Wang \cite{Guo13} demonstrate that filtered images will affect handcrafted features (e.g., SIFT) and degrade the recognition accuracy. Therefore, in this work, we assume CNN-based approaches (e.g., RAPID net) are strong baseline for comparison, and focus on exploring different CNN structures with category information and pairwise aesthetic comparison for filter recommendation.} The results show that even a simple single-column CNN can beat the performance of FV by almost 5\%. Besides, in many other fields, CNN is widely used and outperforms traditional methods. For these reasons, we focus on exploring different CNN structures rather than comparing the performance between non-CNN and CNN-based methods.

\begin{figure}
\centering
\includegraphics[width=3in, keepaspectratio=true]{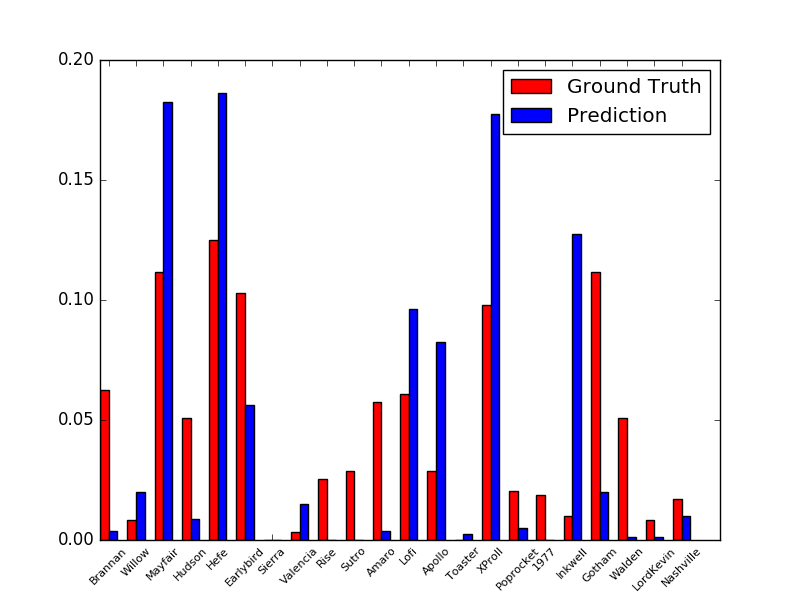}
\caption{Filter preference distribution across image categories.  The chart demonstrates that the distributions of the ground truth and our prediction are similar. Both of them concentrate on certain filters which are commonly selected (e.g., Mayfair, Hefe, XProII). The difference is discussed in Sec. V-C.}
\end{figure}

%-------------------------------------------------------------------------
\section{Conclusions}
In this paper, we present a novel approach that learns aesthetic representations for filter recommendation. Different from traditional image quality studies, the proposed method learns aesthetic responses from pairwise comparison. This idea is achieved by designing a multi-column CNN with a specific loss function. The involvement of image category also improves the performance of filter ranking by multi-task learning. Besides, we introduce a publicly available dataset, Filter Aesthetic Comparison Dataset (FACD), with crowdsourcing labels for pairwise filter comparison. It provides more than 40K filtered image pairs with user preference. To the best of our knowledge, this is the first work that integrates pairwise filter ranking and CNN architectures for image aesthetic response learning. We believe that the newly created FACD and the ideas/observation in this paper can help further research works on image aesthetic analysis. In the future, applying specific structures for different objectives or even integrating additional image information may promote the performance of aesthetic ranking. Meanwhile, the ranking order may be more informative than pairwise comparison; hence, it is also essential to utilize those pair information and design suitable loss functions for aesthetic ranking. Therefore, we will further explore possible ranking methods under CNN structures and compare with traditional approaches.

%\section*{Acknowledgment}

\section*{Acknowledgment}

This work was supported in part by the Ministry of Science and Technology, Taiwan, under Grant MOST 104-2622-8-002-002 and MOST 105-2218-E-002-032, and in part by MediaTek Inc. and grants from NVIDIA and the NVIDIA DGX-1 AI Supercomputer.

%\ifCLASSOPTIONcaptionsoff
%  \newpage
%\fi



\bibliographystyle{IEEEtran}
\bibliography{TMM-cite}


\vspace{-1cm}
\begin{IEEEbiography}
[{\includegraphics[width=1in,height=1.25in,clip,keepaspectratio]{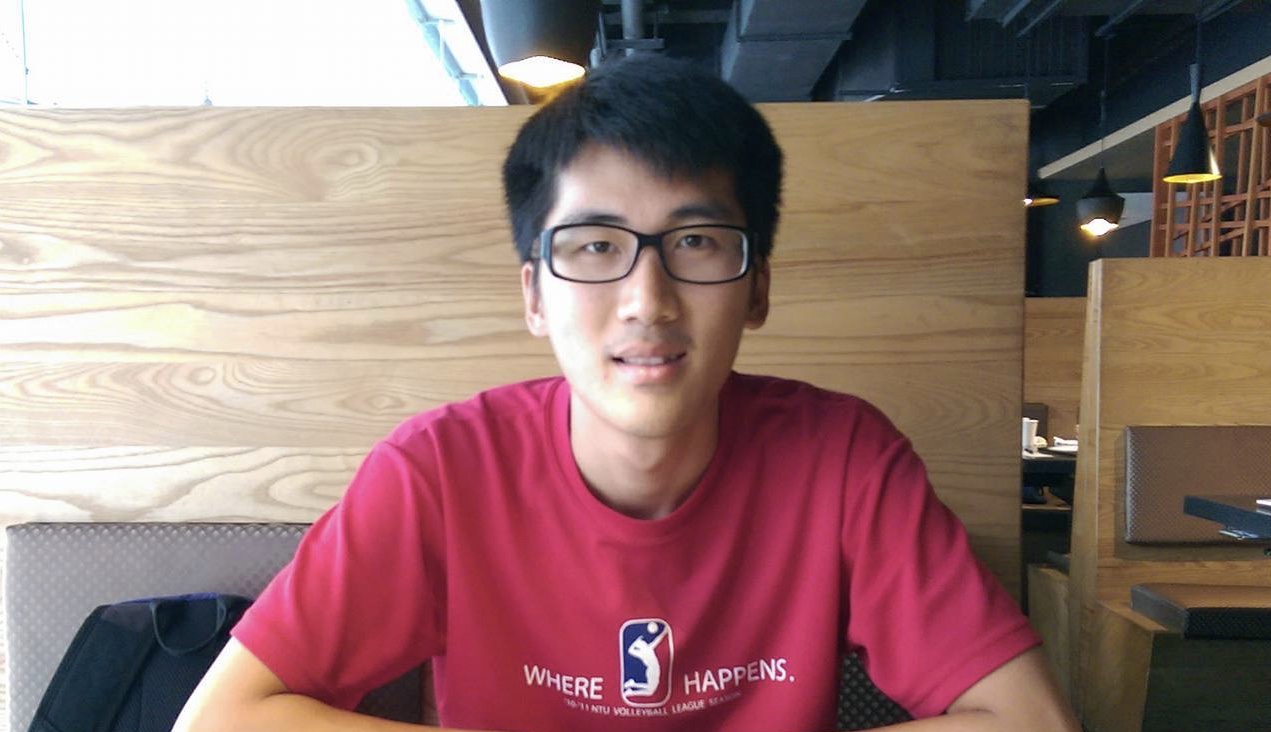}}]
{Wei-Tse Sun}
is fulfilling the mandatory military service in Taiwan. He received the B.S. and M.S. degrees in the Department of Computer Science and Information Engineering, National Taiwan University, Taipei, Taiwan, in 2014 and 2016, respectively.

His research interests include computer vision and image quality assessment. 
\end{IEEEbiography}

\vspace{-1cm}
\begin{IEEEbiography}
[{\includegraphics[width=1in,height=1.25in,clip,keepaspectratio]{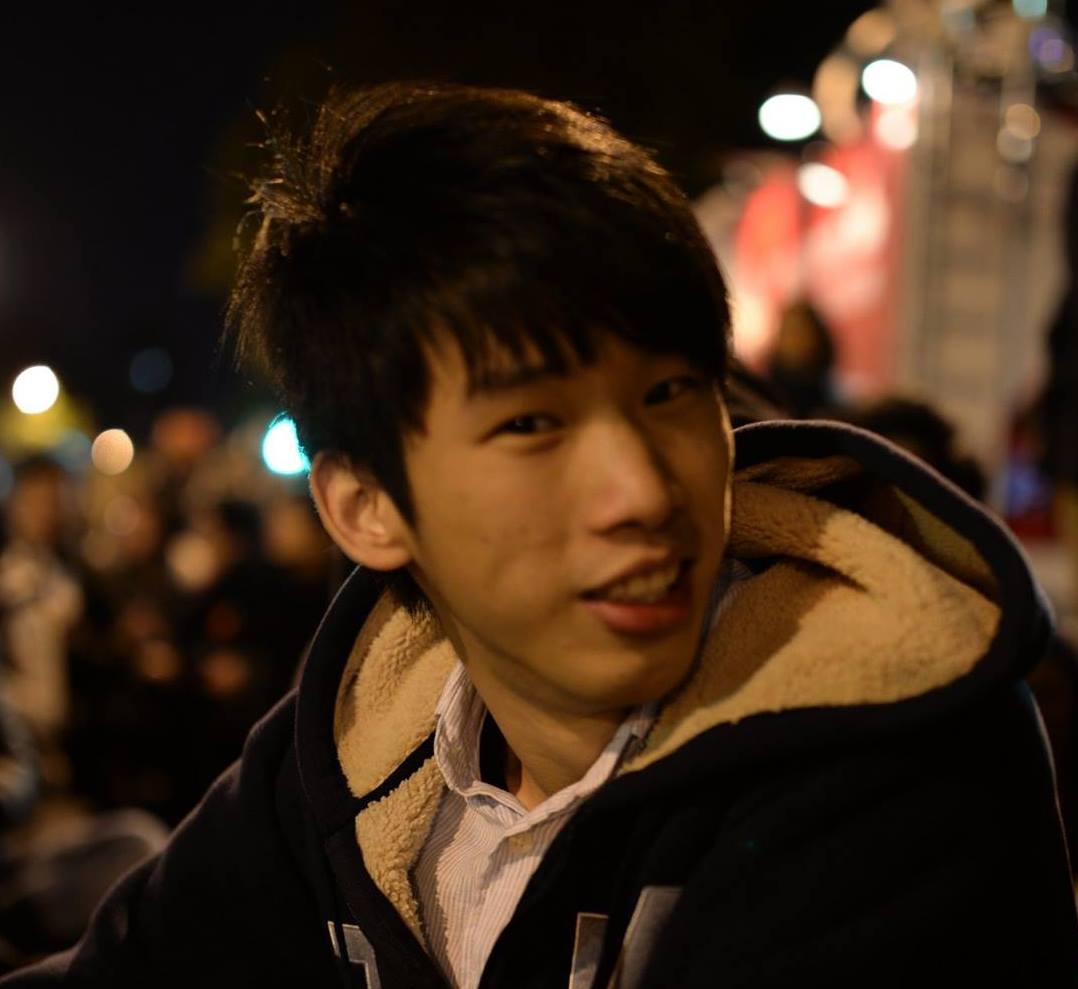}}]
{Ting-Hsuan Chao}
has been worked in Appier Technology Company since 2016. He received the M.S. degree from the Department of Computer Science and Information Engineering, National Taiwan University, Taipei, Taiwan, in 2015. 

His research interests include computer vision and deep learning. 
\end{IEEEbiography}

\vspace{-1cm}
\begin{IEEEbiography}
[{\includegraphics[width=1in,height=1.25in,clip,keepaspectratio]{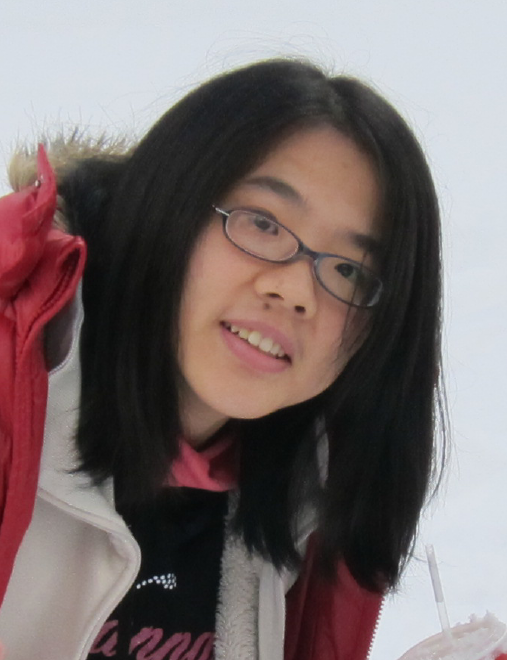}}]
{Yin-Hsi Kuo}
received the M.S. degree in computer science and information engineering from National Taiwan University, Taipei, Taiwan, where she is currently pursing the Ph.D. degree with the Graduate Institute of Networking and Multimedia. 

Her current research interests include multimedia content analysis, image retrieval, and deep learning for multimedia.
\end{IEEEbiography}

\vspace{-1cm}
\begin{IEEEbiography}
[{\includegraphics[width=1in,height=1.25in,clip,keepaspectratio]{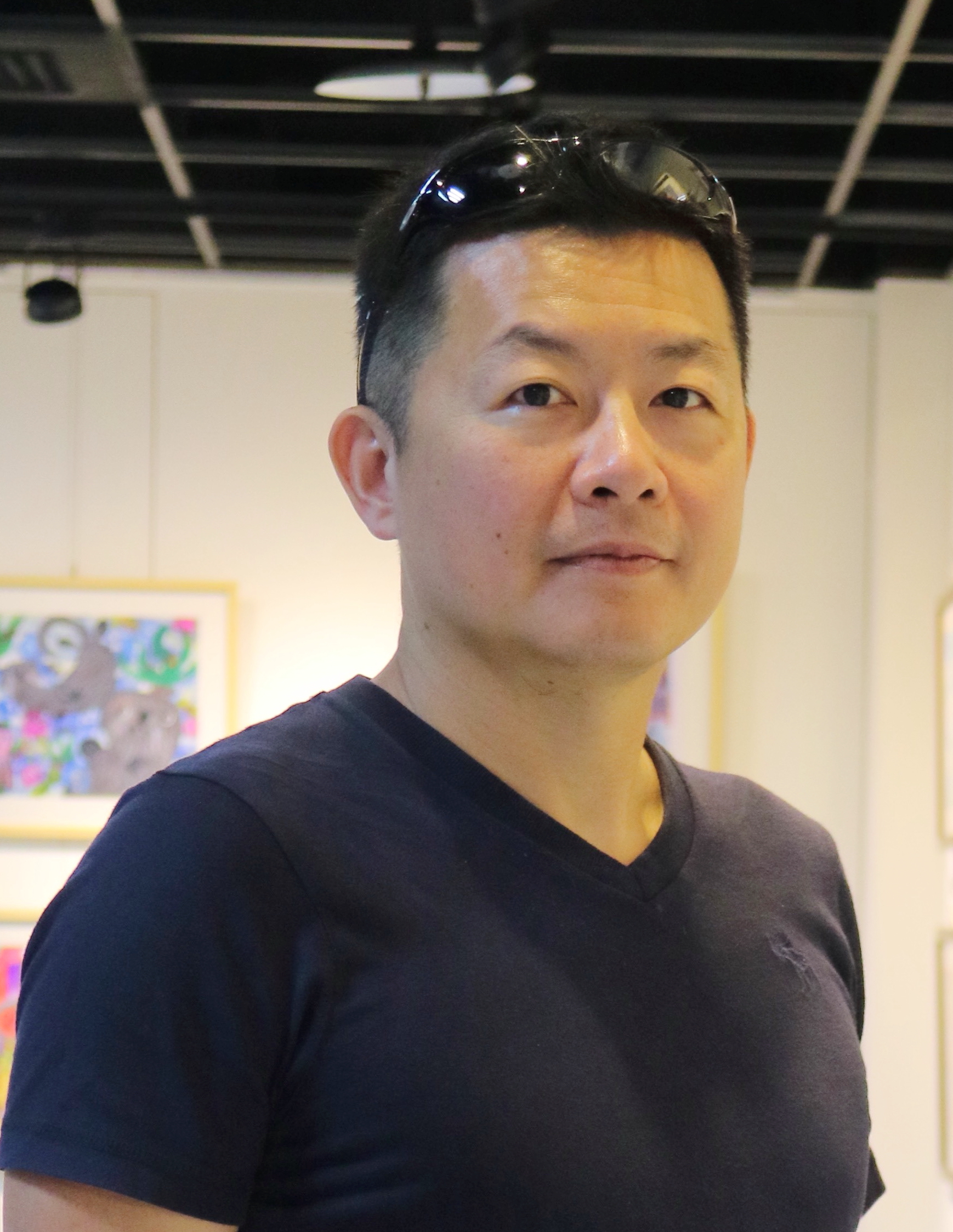}}]{Winston H. Hsu}
(S'03--M'07--SM'12) received the Ph.D. degree in electrical engineering from Columbia University, New York, NY, USA.

He previously worked in the multimedia software industry. Since 2007, he has been a Professor with the Graduate Institute of Networking and Multimedia, National Taiwan University, Taipei, Taiwan. He is also with the Department of Computer Science and Information Engineering, National Taiwan University, Taipei, Taiwan. His research interests include computer vision, machine intelligence, image/video indexing and retrieval, and mining over large-scale databases.

Dr. Hsu serves as the Associate Editor for IEEE Transactions on Multimedia and on the Editorial Board for the IEEE MULTIMEDIA MAGAZINE.
\end{IEEEbiography}


%\vfill


\end{document}